\documentclass[letterpaper]{article} 
\usepackage{aaai24}  
\usepackage{times}  
\usepackage{helvet}  
\usepackage{courier}  
\usepackage[hyphens]{url}  
\usepackage{graphicx} 
\usepackage{natbib}  
\usepackage{caption} 
\usepackage{algorithm}
\usepackage{algorithmic}
\usepackage{xtab,booktabs}
\usepackage{newfloat}
\usepackage{listings}
\DeclareCaptionStyle{ruled}{labelfont=normalfont,labelsep=colon,strut=off} 
\floatstyle{ruled}
\newfloat{listing}{tb}{lst}{}
\floatname{listing}{Listing}
\usepackage{adjustbox}
\usepackage{booktabs}
\usepackage{siunitx}
\usepackage{ragged2e}
\usepackage{threeparttablex}
\usepackage{calc}
\usepackage{etoolbox}
\usepackage{xcolor}

\newcolumntype{R}[1]{>{\RaggedRight\arraybackslash}m{#1}}
\newcolumntype{P}{>{\RaggedRight\arraybackslash}p{(\linewidth-2.2cm-14\tabcolsep)/6}}

\usepackage{comment}
\nocopyright 

\title{MetaHate: A Dataset for Unifying Efforts on Hate Speech Detection}
\author{
    Paloma Piot\textsuperscript{\rm 1},
    Patricia Martín-Rodilla\textsuperscript{\rm 1},
    Javier Parapar\textsuperscript{\rm 1}
}
\affiliations{
    \textsuperscript{\rm 1}
    IRLab, CITIC Research Centre, Universidade da Coruña\\
    paloma.piot@udc.es, patricia.martin.rodilla@udc.es, javier.parapar@udc.es
}

\begin{document}

\maketitle

\begin{abstract}

Hate speech represents a pervasive and detrimental form of online discourse, often manifested through an array of slurs, from hateful tweets to defamatory posts. As such speech proliferates, it connects people globally and poses significant social, psychological, and occasionally physical threats to targeted individuals and communities. Current computational linguistic approaches for tackling this phenomenon rely on labelled social media datasets for training. For unifying efforts, our study advances in the critical need for a comprehensive meta-collection, advocating for an extensive dataset to help counteract this problem effectively. We scrutinized over 60 datasets, selectively integrating those pertinent into MetaHate. This paper offers a detailed examination of existing collections, highlighting their strengths and limitations. Our findings contribute to a deeper understanding of the existing datasets, paving the way for training more robust and adaptable models. These enhanced models are essential for effectively combating the dynamic and complex nature of hate speech in the digital realm.

\end{abstract}

\section{Introduction and Motivation}
\label{sec:intro}

In recent years, the pervasive influence of social media and online platforms has facilitated unprecedented connectivity and communication on a global scale. However, this interconnectedness has also brought to the forefront a concerning rise in the prevalence of hate speech—an issue that transcends geographical boundaries and cultural differences \citep{pewresearch2021onlineharassment, hickey2023auditing}. Different studies report that more than 30\% of young people have been victims of cyber hate by their peers \cite{kansok2023systematic}. As society navigates this digital age, the urgency to address hate speech and its detrimental consequences has never been more critical. For identifying different types of offensive messages, machine learning models were suggested in 1997 \citep{Spertus1997SmokeyAR}. Nowadays, with the rise of Large Language Models (LLMs), the need for vast datasets is crucial \citep{kurrek-etal-2020-towards}. In this paper, we scrutinize state-of-the-art hate speech datasets and assess them to organize a large-scale meta-collection for hate speech detection on social media.

There is no formal definition of hate speech, but previous works \citep{davidsonhatespeech,fountahatespeech,mathewb2020hatexplain,hatelingo2018elsherief,ElSherief2018,silva2016analyzing,hatemm2023} deepened on this topic, defining it as ``\textit{language characterized by offensive, derogatory, humiliating, or insulting discourse \citep{fountahatespeech} that promotes violence, discrimination, or hostility towards individuals or groups \citep{davidsonhatespeech} based on attributes such as race, religion, ethnicity, or gender \citep{hatelingo2018elsherief,ElSherief2018, hatemm2023}}''. Under this definition, which aligns very well with the United Nations one \citep{unhatespeech}, we frame our work by differentiating hate speech from non-hate and offensive speech. 

One of the challenges faced in hate speech detection is the lack of standardized datasets \citep{hatelingo2018elsherief,Poletto2020,toraman-etal-2022-large}, evaluation metrics \citep{rottger-etal-2021-hatecheck}, and benchmark models \citep{Poletto2020}. The motivation for creating a meta-collection lies in the recognition that individual efforts to combat hate speech are essential but often limited in scope. By consolidating diverse datasets, methodologies, and models from various contributors, a meta-collection serves as a centralized resource that empowers researchers and practitioners to collaborate, learn from each other's experiences, and collectively advance the field of hate speech detection. This collaborative approach not only accelerates progress but also ensures the development of more robust and generalizable models that can adapt to the ever-evolving nature of online hate speech. In this paper, we advocate for the necessity of a meta-collection on hate speech detection as a pivotal step towards fostering generalizable machine learning detection models. By harnessing collective knowledge and resources, the meta-collection endeavours propel the field forward, providing a united front against the proliferation of online hate speech in our interconnected world.

\section{Data Acquisition and Preparation}
\label{sec:data}

Over the past few years, numerous efforts have been made to create datasets for hate speech analysis \citep{davidsonhatespeech,Golbeck2017,fountahatespeech}. The community has a widely recognized list that aims to collect all available hate speech corpora: hatespeechdata. While this repository is a valuable source, it only provides a list of various dataset publications and their links. Nevertheless, the website also studied 63 datasets, of which 25 are in English, focusing on the best practices for creating datasets for detecting hate speech \citep{Vidgen2020}. Studies such as \citet{Poletto2020} have concentrated on studying all available corpora resources to detect hate speech. This has resulted in a comprehensive survey that highlights the numerous benchmark datasets available for evaluating abusive language. Nevertheless, none of these studies has released any data.

Our study analyzes over 60 hate speech detection datasets, selecting those relevant to our topic for our meta-collection. To gather data on hate speech, we conducted a thorough exploration in search engines and repositories. Additionally, we undertook a comprehensive review of academic literature, including works authored by \citet{Vidgen2020, Poletto2020, Mody2023}, to enhance the scope and depth of our data collection efforts. All accessed datasets were devoid of personal information, ensuring that users' personally identifiable information remained uncompromised. We meticulously filtered all the examined related works to assess their compatibility with MetaHate. Our selection adheres to specific criteria: (1) inclusion of only social media texts authored by humans, excluding datasets derived from alternative sources or synthetic data; (2) incorporation of datasets aligned with a hate speech definition similar to ours, not considering offensive content as hate speech as it doesn't correspond to our specified hate speech criteria; (3) focusing on English datasets, given the high volume of data available and the inherent coherence within our internally collected information.

We attempted to incorporate additional hate speech datasets that are pertinent to our study, such as the constructed in the \citet{hatelingo2018elsherief} study. Unfortunately, these datasets are not publicly accessible, and our attempts to reach out to the authors went unanswered. Details about these datasets, along with other significant hate speech studies that did not align with our work's definition, such as the research conducted by \citet{hatemm2023}, are compiled in MetaHate website\footnote{\url{https://irlab.org/metahate.html}}. We gathered the data and reported the actual dataset sizes after removing duplicated entries. Table \ref{tab:datasets} summarized the datasets integrated in MetaHate, together with other abusive datasets. Next, we present the analysed datasets.

\textbf{Online Harassment 2017} \citep{Golbeck2017}: Golbeck et al. conducted an exploratory analysis of offensive terms to extract hashtags, lexicon and word structures, which they used to scrape data from Twitter for their dataset. They focused on hate speech and followed a binary classification approach. Contributed: \num{19838} posts.

\textbf{OLID 2019} \citep{zampieri-etal-2019-predicting}: Zampieri et al. compiled a list of keywords and constructions that are often included in offensive content to scrape tweets and create their dataset. They propose three classification levels, and we've used the first one (A) - the binary level of offensive speech. We found that their definition of offensive content and hierarchy was similar to our definition of hate speech. Contributed: \num{14052} posts.

\textbf{HASOC 2019-2021} \citep{Mandl2019, Mandl2020, Modha2021}: HASOC is a track featured at the FIRE conference that creates resources for identifying hate speech. It was first introduced in 2019, where participants were provided with a dataset of tweets classified as hate and no hate. In the following editions, the same task was proposed for hate classification in English. The 2019 dataset is publicly available, but the 2020 and 2021 datasets require a password that we were unable to obtain despite reaching out to the authors. Contributed: \num{6981} posts.

\textbf{A Curated Hate Speech Dataset 2023} \citep{Mody2023}: Mody et al. attempted to construct a comprehensive large dataset. When we initially downloaded their dataset, we discovered \num{842334} posts, but after removing duplicates, we were left with \num{560385} samples. Their published curated dataset doesn't provide many details and lacks any experimental analysis of the data. Their sources are derived from 18 datasets, but some of them are no longer available, and there are instances of duplicated datasets in their list. Contributed: \num{560385}. 

\textbf{Measuring Hate Speech 2020 \& 2022} \citep{kennedy2020rasch, sachdeva-etal-2022-measuring}: Collaborative endeavours ended up in the compilation of a dataset sourced from three diverse platforms: Twitter, Reddit, and YouTube. To assess the hatefulness of the content, the authors opted for a linear hate speech scale, employing Rasch item response theory (IRT). Annotator ratings were transformed into this hate scale, where high values ($>$0.5 approx.) indicated more hateful texts. Values between -1 and 0.5 were assigned to texts perceived as neutral or ambiguous, while those below -1 denoted counter or supportive speech. Contributed: \num{39565} posts.

\textbf{Intervene Hate 2019} \citep{qian-etal-2019-benchmark}: Qian et al. directed their attention to the Reddit and Gab platforms. Their objective was to automatically generate responses for intervention during online conversations containing hate speech. For Reddit, they gathered the top 200 hottest submissions from toxic subreddits and reconstructed the conversation. On Gab, they employed hate keywords to retrieve the original posts and reconstruct the conversation context. Contributed: \num{45170} posts.

\textbf{ETHOS 2022} \citep{Mollas2022}: Mollas et al. meticulously curated two distinct collections, from toxic subreddits and hatebusters: the first comprising \num{998} comments labelled for the presence or absence of hate speech content, and the second consisting of \num{443} hate speech messages categorized through multiclass and multilabel classification. Both datasets were meticulously assembled using data from Reddit and YouTube. Contributed: \num{998} posts.

\textbf{Hate in Online News Media 2018} \citep{Salminen2018}: Salminen et al. meticulously labelled comments extracted from YouTube videos and Facebook posts associated with an online news and media company that maintained a highly active presence on social media platforms. The labeling process involved categorizing comments into hateful and neutral ones, and additionally specifying the target of the hate. Contributed: \num{3214} posts.

\textbf{Supremacist 2018} \citep{de-gibert-etal-2018-hate}: The authors randomly sampled posts from Stormfront, a neo-Nazi Internet forum, as scraping data from this forum, had an intrinsic nature of being hateful, and labelled the data on hate and non-hate speech. Contributed: \num{10534} posts.

\textbf{The Gab Hate Corpus 2022} \citep{Kennedy2022}: In an effort to mitigate potential biases introduced by keyword-based strategies, Kennedy et al. opted to build a hate speech dataset from a less conventional social network, Gab. Gab is known as a right-wing social platform where hate or abusive comments are more prevalent, and the researchers chose to randomly sample diverse publications and classify them as an assault on human dignity or not. Contributed: \num{27434} posts.

\textbf{HateComments 2023} \citep{guptahatevideos}: Gupta et al. have recently created a hate speech dataset from YouTube and BitChute video comments. The comments were tagged on a binary level and the dataset also includes some video context. Contributed: \num{2070} posts.

\textbf{Toxic Spans 2021} \citep{pavlopoulos-etal-2021-semeval}: Various initiatives addressing the issue of hate speech were undertaken within the framework of the ``SemEval-2021'' task. The primary goal was to anticipate the specific spans within posts that contributed to their classification as toxic. To accomplish this, a dataset sourced from Civil Comments was meticulously curated and labelled with spans to indicate the precise portions responsible for the toxic labels. Contributed: \num{10621} posts.

\textbf{Ex Machina 2016} \citep{exmachina2016}: In their early endeavours, Wulczyn et al. dedicated their efforts to constructing a sizable dataset by randomly sampling posts and incorporating comments from blocked accounts. Their collection comprises over \num{100000} Wikipedia comments, focusing on personal attacks on a binary level. Contributed: \num{115705} posts.

\textbf{Context Toxicity 2020} \citep{pavlopoulos2020context}: Pavlopoulos et al. delved beyond individual messages, investigating the impact of context on hate speech detection. Their research unfolded in two parts: firstly, they constructed a small dataset of \num{250} comments from Wikipedia to assess how annotation outcomes were influenced by the provision of context. Secondly, they expanded their dataset to include nearly \num{20000} Wikipedia comments. Half of the comments were annotated without context, and the remaining half were annotated with context, on toxic or non-toxic speech. Contributed: \num{19842} posts.

\textbf{BullyDetect 2018} \citep{BinAbdurRakib2018}: Bin Abdur Rakib and Soon conducted research on cyberbullying detection by building a binary corpus of posts from Reddit. However, it is not clear how they created the dataset. Their main focus was on training a word embedding model to then train a Random Forest Classifier in order to evaluate their dataset. Contribution: \num{6562} posts.

\textbf{US 2020 Elections} \citep{grimminger-klinger-2021-hate}: Grimminger and Klinger directed their hate detection efforts toward the political domain, specifically targeting the 2020 United States Elections. They sampled data from Twitter using search terms related to mentions of presidential candidates, various hashtags reflecting voter alignment, campaign slogans, and even nicknames of the candidates. In their comprehensive analysis, the dataset was meticulously annotated following a binary classification. Contributed: \num{2999} posts.

\textbf{``Call me sexist but'' 2021} \citep{Samory2021}: The ``Call me sexist, but...'' dataset was meticulously derived from an extensive pool of over \num{13000} tweets. These texts underwent a detailed binary classification process, discerning between instances of sexism and those devoid of such content. Four distinct creation approaches were deployed, including (1) the utilization of sexism scales, (2) incorporating \citep{Waseem2016}'s sexism tweets, (3) integrating benevolent sexism tweets from \citep{jha-mamidi-2017-compliment}, and (4) collecting tweets containing the phrase ``call me sexist, but''. Contributed: \num{3058} posts.

\textbf{Hateval 2019} \citep{basile-etal-2019-semeval}: SemEval is a series of workshops that focus on the evaluation and comparison of computational semantic analysis systems. They provided the participants with a dataset, of which the construction strategy focused mainly on fetching tweets by keywords and monitoring haters and victims. In 2019, task 5 topic was hate speech, where the first task was about a binary classification task. Contributed: \num{12747} posts.

\textbf{Hate Offensive 2017} \citep{davidsonhatespeech}: This dataset, curated from hate lexicon, consists of \num{24783} entries from Twitter, which have been meticulously categorized into three distinct classes, namely hate speech, offensive language, and neutral language. Contributed: \num{24783} posts.

\textbf{TRAC1 2018} \citep{kumar-etal-2018-aggression}: Kumar et al. collaborate with a multilingual dataset, including samples from Twitter and Facebook in English. They fetched data from more than 40 pages discussing controversial topics on Facebook and used popular hashtags around different topics to retrieve data from Twitter. They annotated the data into three different levels of aggressiveness. Contributed: \num{14587} posts.

\textbf{ENCASE 2018} \citep{fountahatespeech}: This dataset was built by collecting random samples of tweets and boosting a sample that represents 12.5\% of the dataset. This sample shows a strong negative polarity and contains at least one offensive word from hate speech lexicons. The tweets were tagged in one of these four different categories: abusive, normal, spam and hateful. Contributed: \num{91950} posts.

\textbf{MLMA 2019} \citep{ousidhoum-etal-2019-multilingual}: This multilingual work, which includes English posts from Twitter, shares that the authors were able to build a multilabel and multiclass dataset of \num{5593} entries by looking to tweets that contained keywords and phrases related to hate. Contributed: \num{5593} posts.

\textbf{HateXplain 2020} \citep{mathewb2020hatexplain}: Mathew et al. directed their attention towards examining the bias and interpretability facets of hate speech. They compiled a dataset from Twitter by extracting content using a lexicon and incorporating the \citet{Mathew2019} dataset sourced from the Gab platform. Each post was evaluated by three annotators who assigned labels based on three distinct categories: hate, offensive, and normal. Contributed: \num{20109} posts.

\textbf{Hateful Tweets 2022} \citep{Albanyan2022}: This research aimed to construct a corpus for analyzing the context and counter-narratives of hate speech texts. Utilizing works by \citet{Waseem2016}, \citet{fountahatespeech}, and \citet{davidsonhatespeech}, the authors selectively curated entries featuring racist, sexist, abusive, hateful, and offensive content. Subsequently, they sought to retrieve replies and contextual information for these targeted tweets. Contributed: \num{1141} posts.

\textbf{Multiclass Hate Speech 2022} \citep{toraman-etal-2022-large}: The authors created this dataset by using a diverse set of keywords and hashtags spanning various topics, including religion, gender, racism, politics, and sports. They categorized the tweets into hate, offensive, and normal texts. Contributed: \num{68597} posts.

\textbf{Slur Corpus 2020} \citep{kurrek-etal-2020-towards}: Almost \num{40000} posts sourced from Reddit were sampled using three slurs targeting discrimination across sexuality, ethnicity, and gender, from the Pushshift Reddit dataset \citep{baumgartner2020-pushshift}. Kurrek et al. established a comprehensive taxonomy and categorized their dataset into derogatory, non-derogatory, homonym, appropriation, and noise labels. Contributed: \num{39960} posts.

\textbf{TRAC2 2020} \citep{trac2-dataset}: In endeavours related to TRAC shared tasks, Bhattacharya et al. constructed a dataset of comments extracted from YouTube videos. Adhering to a methodology akin to that of \citet{kumar-etal-2018-aggression}, they distinguished between three types of aggression. Contributed: \num{5329} posts.

\textbf{CAD 2021} \citep{vidgen-etal-2021-introducing}: Introducing a taxonomy comprising six conceptually distinct categories, Vidgen et al. undertook a sampling of posts from various subreddits anticipated to contain abusive content. The texts were meticulously labelled across these categories (directed abuse, counter-speech, etc.). Contributed: \num{23060} posts.

\textbf{Hate Speech A 2016} \citep{Waseem2016}: Waseem and Hovy built this dataset by manually searching on Twitter for common slurs, terms, and phrases related to religious, sexual, gender, and ethnic minorities. They categorized the tweets into three groups: racism, sexism, and none. Contributed: \num{16849} posts.

\textbf{Hate Speech B 2016} \citep{waseem-2016-racist}: Waseem's subsequent works continued to centre around hate speech detection, with a particular emphasis on refining the annotation aspect of this task. In this endeavour, they meticulously annotated almost \num{7000} tweets across four distinct categories—expanding by one category compared to their previous work. The new categories included racism, sexism, both, and none. Contributed: \num{6909} posts.

\textbf{\#MeTooMA 2020} \citep{Gautam2020}: This paper focuses on a specific type of hate speech, about the topic \#MeToo. They narrowed their search where the \#MeToo movement and identified keywords and phrases to create a 75-keyword lexicon and then queried tweets from these countries using this lexicon. The entries were labelled on hate speech, sarcasm, allegation, justification, refutation, support and opposition. Contributed: \num{9889} posts.

\textbf{Hate Speech Data 2017} \citep{mondal2017}: In their study, Mondal et al. incorporated data from both Whisper and Twitter, curating a dataset comprising \num{28309} entries—\num{6157} from Whisper. Distinguishing themselves from conventional approaches, their method involved devising diverse hate speech sentence structures aimed at isolating purely hate speech comments. These structures were then enriched with a hate lexicon, enabling the identification of hate based on various categories (race, class, gender, ethnicity, disability, etc). Another study by \citet{silva2016analyzing} analyzed this dataset to explore the targets of online hate speech. Contributed: \num{6157} posts from Whisper (we were unable to access the Twitter data).

Several other studies on hate speech detection in social media have shifted their focus to low-resource languages and languages beyond English. Noteworthy efforts have been observed in languages like Spanish \citep{Fersini2018,basile-etal-2019-semeval,s19214654}, Portuguese \citep{fortuna-etal-2019-hierarchically}, Italian \citep{sanguinetti-etal-2018-italian,evalita2020}, French and Arabic \citep{ousidhoum-etal-2019-multilingual}, Turkish \citep{toraman-etal-2022-large}, Slovene \citep{10.1007/978-3-030-27947-9_9}, and many more. While our work is centred on constructing a meta-collection in English due to space constraints, we acknowledge the importance of exploring other languages. Future works from our team will delve into addressing this gap. 

Furthermore, we were able to find different hate speech datasets that were not linked to any publication but were uploaded to different platforms like Kaggle. Table \ref{tab:datasets_repos} shows the summary of these datasets.

\section{Meta-Collection Overview}
\label{sec:metahate}

\textbf{MetaHate} represents a comprehensive compilation of recent advancements in hate speech datasets, with a dual focus: (1) detecting hate speech, toxic behaviour, cyberbullying, aggression, and related terminologies, all falling under the umbrella of the defined harmful online content, and (2) analyzing text authored by humans across various social media platforms. Our meticulous review encompassed over 60 datasets dedicated to hate speech detection, ultimately incorporating 36 datasets into our meta-collection. The compilation resulted in \num{1667496} entries, streamlined to \num{1226202} non-duplicated comments. In shaping the scope of our dataset, we prioritized social media comments from platforms like Twitter or Facebook, while excluding synthetic data and sources outside the realm of social media, such as news comments or video game chats.

In an effort to streamline the dataset for broader applicability, we adopted a binary classification: hate or no hate. The primary rationale for this choice is that approximately half of the base datasets use a binary classification approach, while the remaining half adopt a multiclass approach. It is relatively straightforward to convert from a multiclass classification type to a binary one. However, the reverse process would be time-consuming and demanding. We are committed to fostering collaborative research by making our meta-collection available\footnote{\url{https://irlab.org/metahate.html}} to the research community.

\subsection{Dataset analysis}
\label{sec:analysis}

As previously mentioned, MetaHate encompasses \num{1226202} comments gathered from various social media networks: Twitter, Facebook, Reddit, Stormfront, Gab, Whisper, Wikipedia, Civil Comments, YouTube, and BitChute. The strategies employed for the datasets creation include: (1) utilizing lexicons, keywords, hashtags, and phrase structures, and (2) randomly sampling from sites with a likelihood of containing hate content. In terms of conceptualization, the majority of works adopt a binary strategy. However, a vast of them take a multiclass approach, distinguishing between abusive, hate, offensive, or normal speech, among other terms. A few efforts explore a probabilistic approach, assigning a numerical value between 0 and 1 to gauge the degree of hatefulness in a comment. Additionally, some studies opt for a multiclass and multilabel approach and others go a step further, attempting to extract the specific span that contains hate speech within a larger sentence.

Among these entries, \num{253145} instances are labelled as hate, while the remaining \num{973057} are identified as non-hateful. This signifies that 20.64\% of our dataset entries are categorized as hate comments. This observation aligns with the consensus of some existing works, some of them emphasize that while hate speech is a genuine concern online, the overwhelming majority of posts fall within the non-hateful category \citep{Waseem2016}. The average post length is \num{261.80} characters and \num{43.39} words, with the longest post containing \num{2832} words. While this aligns with the character limit of \num{280} in a tweet, our dataset includes posts from platforms like Reddit, where the maximum length can reach \num{40000} characters.

\begin{figure}[htbp]
\centering
\includegraphics[width=\linewidth]{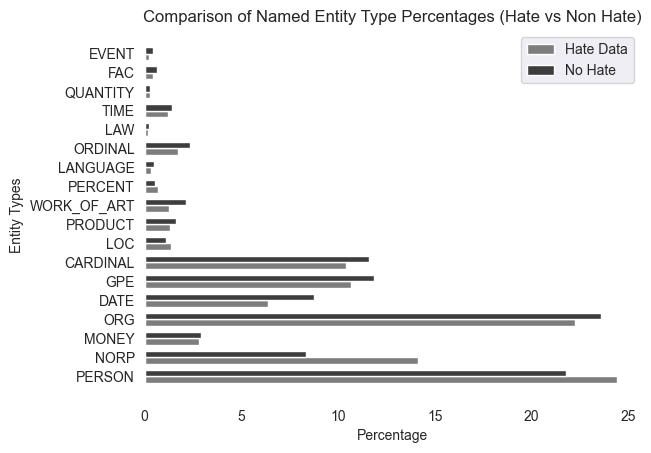}
\caption{Comparison of Named Entity Type percentages between hate and non-hate posts.}

\label{fig:ner}
\end{figure}

\textbf{Lexical Analysis}: First, we conducted a simple term frequency analysis. We observed that within the 20 most frequent keywords in our posts, terms such as \textit{f*ck}, \textit{user}, \textit{people}, \textit{n*gger}, \textit{f*ggot}, \textit{b*tch}, \textit{hate}, \textit{article}, and \textit{woman} emerged. While certain words like \textit{f*ggot} and \textit{b*tch} could be associated with hate speech, others like \textit{article} seem more neutral. Terms such as \textit{people}, \textit{f*ck}, and \textit{women} could be present in both hate and non-hate posts. Additionally, the term \textit{user} is common, as some of our dataset sources replaced user mentions with the word \texttt{user} \citep{zampieri-etal-2019-predicting,ousidhoum-etal-2019-multilingual,mathewb2020hatexplain}. Additionally, in Figure \ref{fig:ner}, we produced a Named Entity Recognition (NER) analysis. When comparing hate and non-hate posts, it's notable that hate comments often include a substantial percentage of references to \textbf{individuals} (almost 25\%), followed by more than a 14\% mentioning \textbf{nationalities, and religious or political groups} (NORP). This suggests a discernible pattern where attacks are frequently targeted at individuals and extensively involve references to nationality, religion, or political affiliation. In contrast, non-hate posts contain fewer references to \textbf{persons}, although they still make up more than 21\%. References to \textbf{nationalities or religious or political} groups hover around 8\%, and we observe that non-hateful posts include more references to \textbf{dates}, \textbf{companies, agencies, institutions} and \textbf{geopolitical entities} (i.e. countries, cities, states).

As a third experiment, we explore the realm of hate and non-hateful data employing topic modelling techniques. The focal point of our investigation was the optimization of hyperparameters, namely (1) the number of topics, (2) the alpha parameter, and (3) the beta parameter, within the LDA topic model. We aimed to maximize topic coherence, ensuring that the identified topics encapsulate meaningful and coherent thematic elements. As a result, we uncovered eight prominent topics within each category (hate and non-hate), each offering valuable insights into distinct thematic elements present in the dataset. For the purpose of this paper, we focus on presenting six out of the eight identified topics.

We examined the different topics through (1) word clouds, showcasing the top 10 words in each topic, providing a qualitative glimpse into the most frequent terms associated with each theme; (2) word count and importance metrics for these top words, offering quantitative insights, allowing us to discern the significance of each term within its respective topic; (3) distribution of document word counts by topic, providing a comprehensive overview of the length and complexity of discussions within each identified theme. Moreover, we employed T-distributed Stochastic Neighbor Embedding (t-SNE) \citep{JMLR:v9:vandermaaten08a} clustering, where we visually represented the relationships and proximity of documents within each of the six hate topics and the six non-hate topics. This technique allowed us to explore the underlying patterns and similarities between documents, providing a more nuanced understanding of the structural dynamics within the hate data.

\begin{figure}[!htbp]
    \centering
    \includegraphics[width=0.8\linewidth]{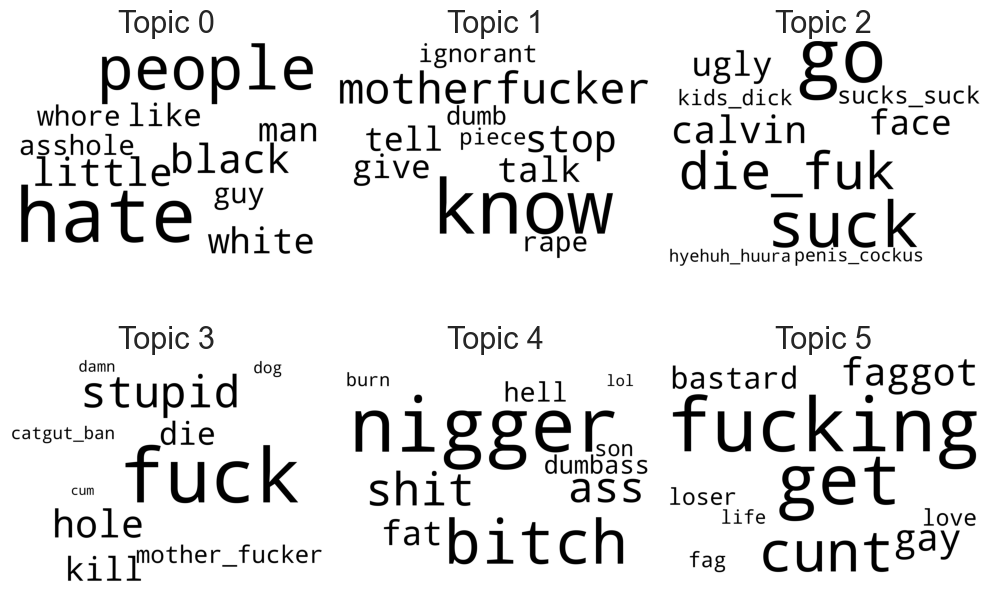}
    \includegraphics[width=0.8\linewidth]{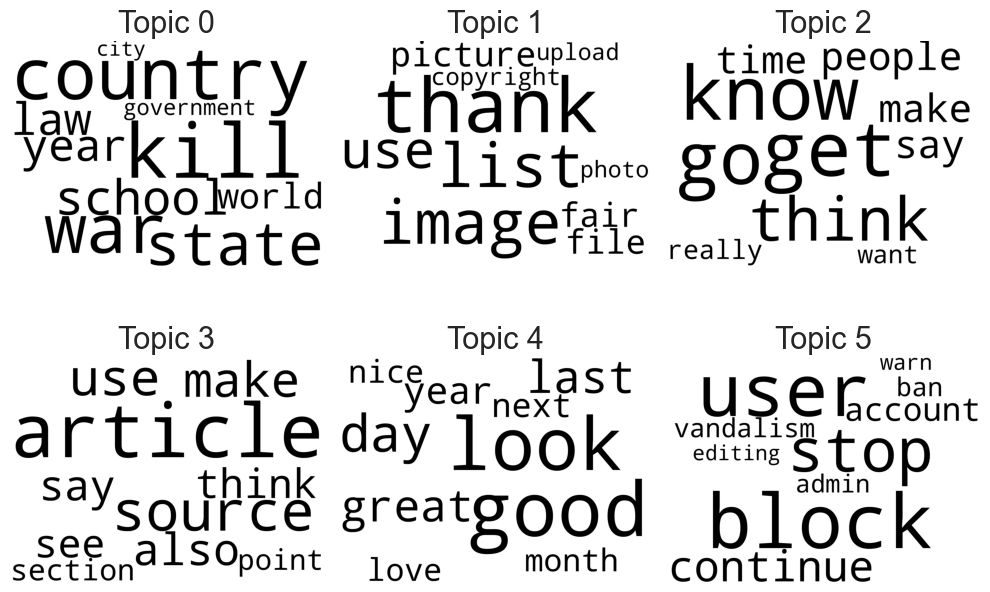}
    \caption{Word clouds of hate topics (up) and non-hate topics (down).}
    \label{fig:topic_wordclouds}
\end{figure}

In Figure \ref{fig:topic_wordclouds}, we present the word clouds showcasing the top 10 words for each of the six topics, distinguishing between hateful posts and non-hateful comments. Notably, the word clouds reveal distinctive lexical patterns characterizing each category. Hate topics prominently feature words such as \textit{hate}, \textit{f*ggot} and \textit{bastard}, indicating explicit negativity and derogatory language. In contrast, the vocabulary associated with non-hateful posts includes terms like \textit{article}, \textit{good}, \textit{think}, \textit{country}, and \textit{image}, reflecting a more positive and constructive linguistic orientation. This divergence in vocabulary underscores the semantic contrast between hate and non-hate content, revealing the distinct linguistic markers that contribute to the characterization of each category.

Furthermore, it is evident that certain topics within the analyzed content are associated with various forms of hate speech. In Topic 0, the presence of words such as \textit{black}, \textit{white}, and \textit{people} is indicative of content aligned with racism themes. Topic 2 features terms like \textit{kids d*ck} and \textit{suck}, suggesting elements of child abuse. Similar associations can be observed in Topics 3 and 4: the former includes expressions like \textit{kill}, \textit{die} and \textit{hole}, pointing to suicide, while the latter hints at fatphobia through the inclusion of words such as \textit{fat} and \textit{b*tch}. Topic 5 top words include \textit{f*ggot}, \textit{gay} and \textit{loser}, inciting hatred of homophobia.

\begin{figure}[!htbp]
    \centering
    \includegraphics[scale=0.28]{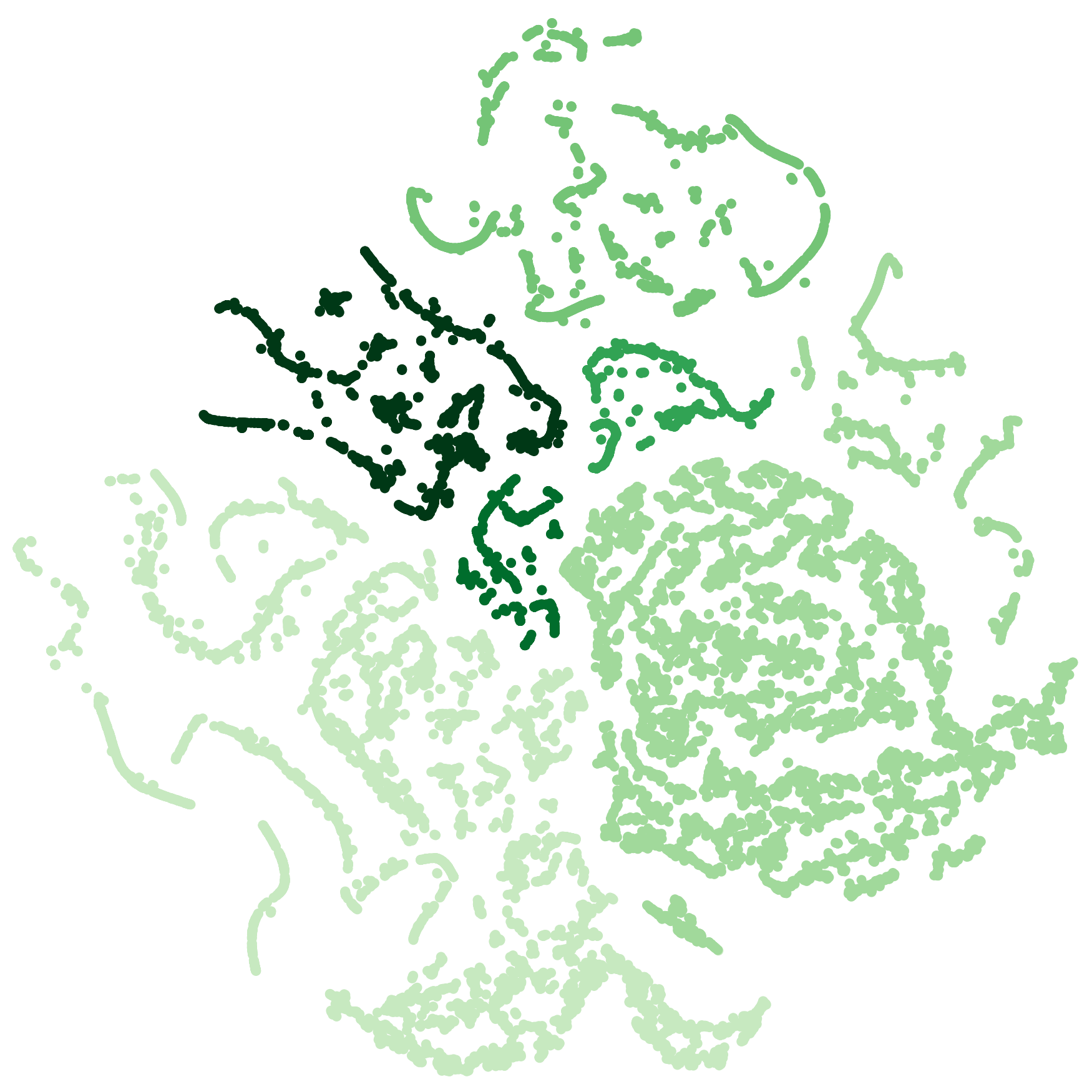}
    \includegraphics[scale=0.28]{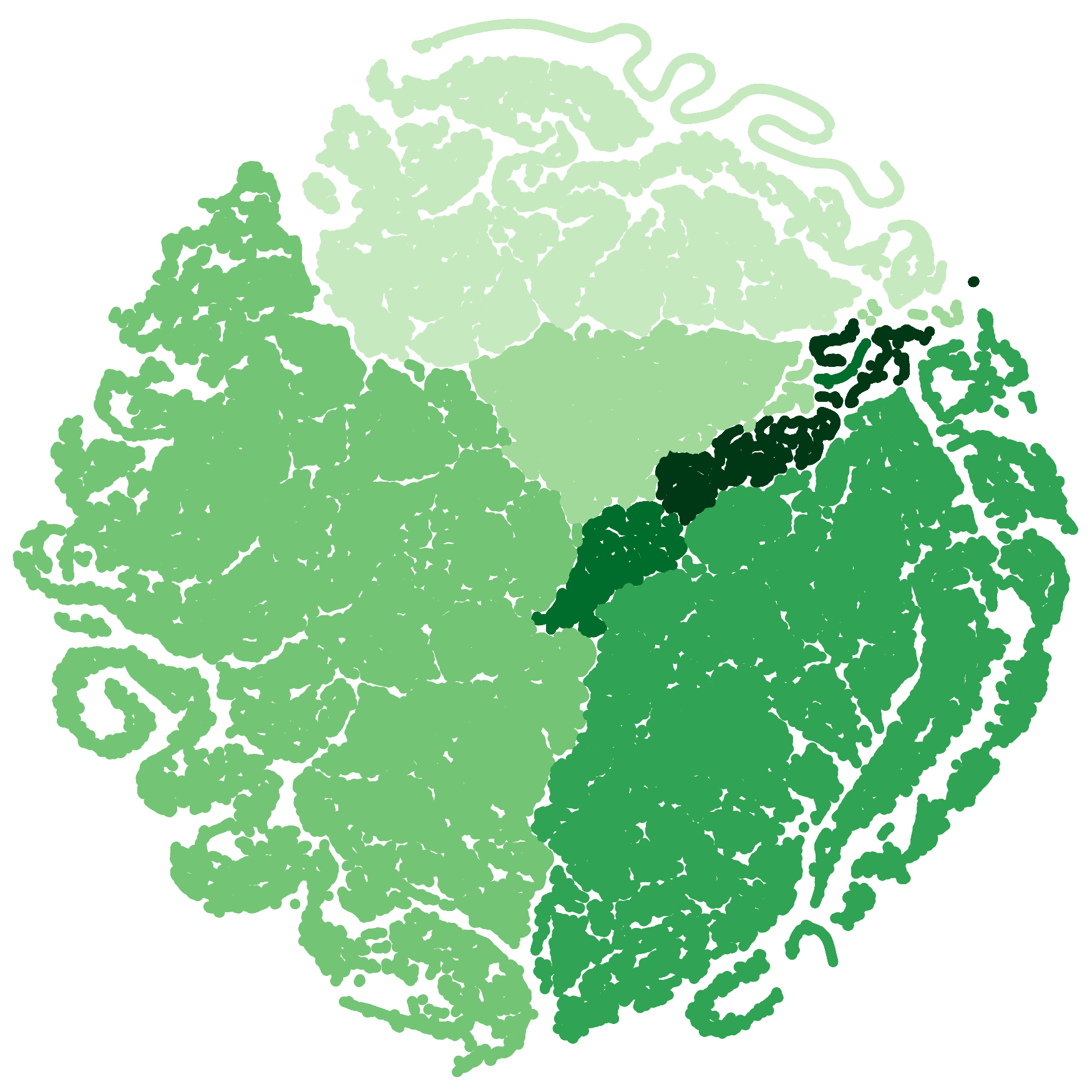}
    \caption{t-SNE diagram illustrating the clustering of hate speech (left) and non-hate speech (right).}
    \label{fig:tsne}
\end{figure}

t-SNE is a dimensionality reduction algorithm used to visualize high-dimensional data. The diagram in Figure \ref{fig:tsne} (left) effectively illustrates the clustering of hate speech from social media LDA topics. Each distinct colour indicates a unique cluster of hate speech, the closer two topics are in the visualization, the more similar they are semantically. We use the topic weights to ensure that the most important topics are given more weight in the visualization. The cluster on the right (topic 1), being the largest, represents the most prevalent hate speech type on social media (including words like \textit{rape}, \textit{motherf*cker} and \textit{ignorant}). The cluster on the left (topic 0) denotes the second most common hate speech, followed by the cluster on the top (topic 2), and so forth. The uneven distribution of the clusters suggests a hierarchical structure of hate speech, with some types being more common than others. While the clusters are somehow separated, they are not randomly distributed, indicating some degree of overlap between them. The t-SNE visualization effectively highlights the semantic relationships between different topics. Topics 1 (right) and 3 (top centre), for instance, share the common term \textit{motherf*cker} and are positioned adjacent to each other, suggesting a close thematic connection. Similarly, topics 0 (left) and 4 (bottom centre), which include the derogatory terms \textit{wh*re} and \textit{b*tch} respectively, exhibit a discernible link in the visualized space.

On the other hand, for non-hateful texts, the t-SNE diagram can be seen in Figure \ref{fig:tsne} (right). The graph shows that the different types of non-hateful content are relatively well-separated, but not as well-separated as the clusters of hate topics. This could be because hate speech is more diverse than non-hate speech. The cluster on the left (topic 2) is the largest cluster, which suggests that it represents the most common type of non-hateful content on social media (including words such as \textit{really}, \textit{think}, \textit{people} and \textit{know}). As the different clusters are not completely separate, this suggests that there is some overlap between the different types of non-hateful content. For example, topics 2 and 3 (left and right), are adjacent as topic 2 terms are quite general and can be related to communication or decision-making, and topic 3 terms are associated with topics related to reading or writing and could be linked to discussions about articles, argumentation or any form of written communication.

\begin{figure}[!htbp]
\centering
\includegraphics[width=0.85\linewidth]{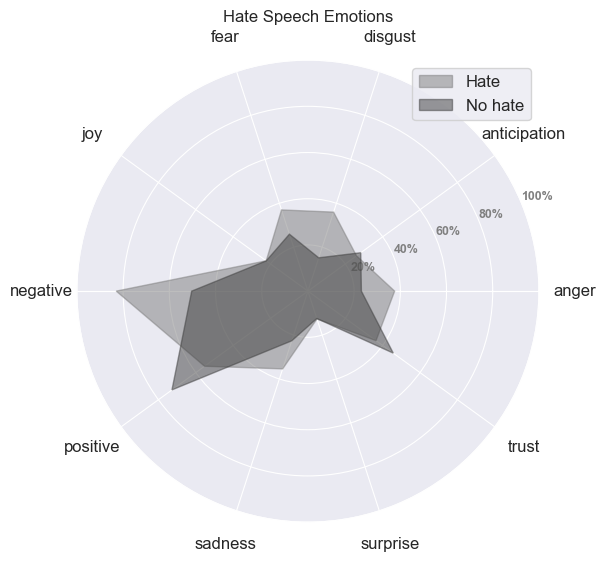}
\caption{Radar plot showing the percentage of posts that contain a word associated with the Plutchick emotions for hate and non-hate data.}
\label{fig:radar}
\end{figure}

\textbf{Psycholinguistic Analysis}: We employed the Plutchik set of emotions \citep{plutchik1980general}, encompassing eight primary emotions (anger, fear, sadness, disgust, surprise, anticipation, trust, and joy) and two basic sentiments (positive and negative). To quantify emotion levels, we utilized the NRC emotion lexicon \citep{Mohammad13}, which comprises words linked to the Plutchik emotions. In our analysis, we computed the percentage of posts within each group (hate, no hate) that included at least one word associated with the primary emotions and sentiments. When comparing the results exposed in Figure \ref{fig:radar}, it is evident that \textbf{negative} emotions are strongly correlated with hate speech posts, while \textbf{positive} emotions are more prevalent in non-hateful content. Additionally, \textbf{fear}, \textbf{disgust}, and \textbf{sadness} exhibit a higher frequency in hate messages than in regular ones. Conversely, emotions like \textbf{trust} are associated with non-hate posts. The predominant emotion in our collection leans towards the negative spectrum, possibly attributed to the construction of hate datasets using keywords and lexicons containing negative terms. Furthermore, there is a significant contrast in the expression of emotions like \textbf{disgust} and \textbf{anger} between hate and non-hate publications.

\begin{figure}[!htbp]
    \centering
    \includegraphics[scale=0.27]{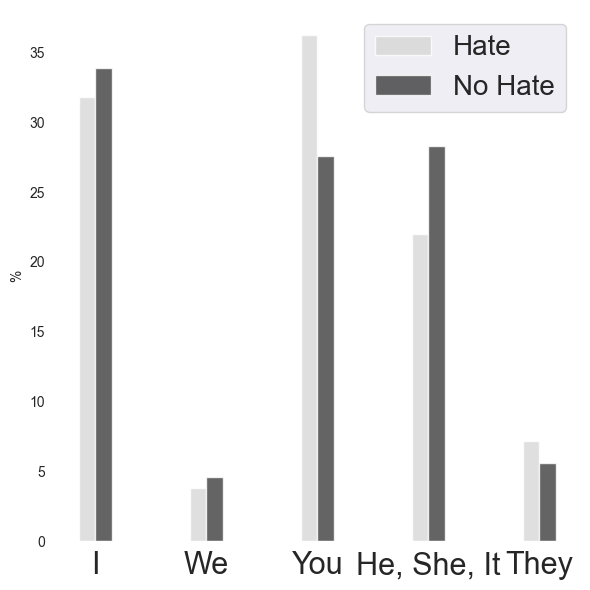}
    \includegraphics[scale=0.27]{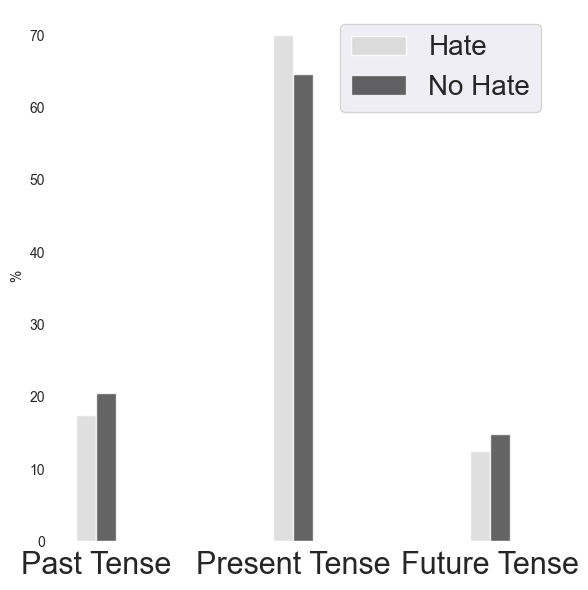}
    \caption{Distribution of pronouns in hate and non-hate posts (left). Hate speech tends to mention \texttt{YOU} more, possibly indicating attacks. On the other hand, non-hate speech more commonly includes references to \texttt{I} and \texttt{HE}, \texttt{SHE}, \texttt{IT}. And distribution of verb tenses in hate and non-hate speech posts (right). Note that \texttt{PRESENT} tenses are more prevalent in hate posts, suggesting possible attacks.}
    \label{fig:verbs-pronouns}
\end{figure}

Additionally, as we can see in Figure \ref{fig:verbs-pronouns} (left), we observe a higher prevalence of second-person pronouns in hate speech posts (36.23\%), while first-person singular and third-person singular pronouns are more common in non-hate posts (33.87\% and 28.31\%, respectively). Moreover, as reflected in Figure \ref{fig:verbs-pronouns} (right), hate speech posts predominantly use the present tense (70.01\%), with 17.50\% in the past tense. In contrast, non-hate posts favour the present tense (64.67\%), with 20.49\% in the past tense. This implies that, although present tenses are predominant in both types of posts, hate speech comments exhibit a higher proportion of present tenses.

\section{Baselines}
\label{sec:baselines}

In this section, we present our efforts to establish a common baseline for hate speech detection on social media. Our objective is to introduce basic text classification methods that can serve as foundational benchmarks for upcoming hate speech classification experiments. To achieve this, we've opted for conventional models like the Support Vector Machine (SVM), utilizing TF-IDF features for text features. Additionally, inspired by recent works \citep{mathewb2020hatexplain,Samory2021,Kennedy2022}, we explored the inclusion of CNNs and BERT in our experimental framework. In all our experiments, the dataset was divided into training and test sets, with a test size of 20\%.

\textbf{Support Vector Machine (SVM)}: We employed a linear SVM model with a feature limit set at 1M, excluding stop words, and implementing TF-IDF unigrams. 

\textbf{Convolutional Neural Network (CNN)}: We established a neural network and conducted training for one epoch on our tokenized sentences, ensuring padding to a maximum length of 512 tokens. The network architecture consists of an embedding layer, flattening, a dense layer with ReLU activation, and a final dense layer with a sigmoid activation for binary classification. The model was compiled using binary cross-entropy loss and the Adam optimizer.

\textbf{BERT}: We used the BERT base uncased model for text classification by loading the pre-trained implementation from the HuggingFace library. We did not do any additional hyperparameter tuning, but we used a learning rate of 5\textit{e}\textsuperscript{-5}, a sequence length of 512 during 3 epochs and a batch size of 32.

\begin{table}[H]
\centering
\footnotesize
\begin{tabular}{l rrrr}
\toprule
\textbf{Method} & \textbf{ACC} & \textbf{F1} & \textbf{F1\textsubscript{MICRO}} & \textbf{F1\textsubscript{MACRO}}\\
\midrule
SVM & 0.85 & 0.84 & 0.85 & 0.74 \\
\midrule
CNN & 0.86 & 0.84 & 0.86 & 0.74 \\
\midrule
BERT & \textbf{0.89} & \textbf{0.88} & \textbf{0.89} & \textbf{0.80} \\
\bottomrule
\end{tabular}
\caption{Hate speech detection results run on MetaHate.}
\label{tab:results}
\end{table}

The results, as shown in Table \ref{tab:results}, reveal notable performance variations among the models. The traditional SVM and CNN models demonstrated competitive accuracies of 0.85 and 0.86, respectively, with corresponding F1 scores of 0.84. In contrast, the \textbf{BERT} model outperformed both, achieving an accuracy of \textbf{0.89} and higher F1 scores of 0.88, 0.89 (micro), and 0.80 (macro). This suggests that BERT, as a pre-trained language model, exhibits superior discriminatory power in capturing intricate patterns within hate speech data, leading to enhanced classification accuracy and nuanced performance across micro and macro F1 metrics. This outcome serves as a robust benchmark, showcasing the model's effectiveness in discerning hate speech within English-language posts on social media platforms using the MetaHate dataset.

\begin{figure}[!htbp]
    \centering
    \includegraphics[scale=0.29]{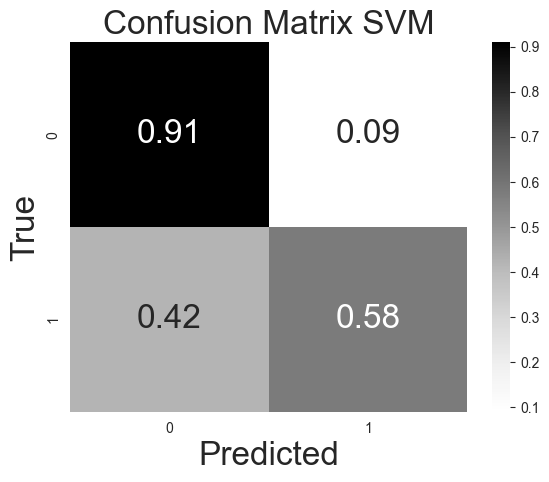}
    \includegraphics[scale=0.29]{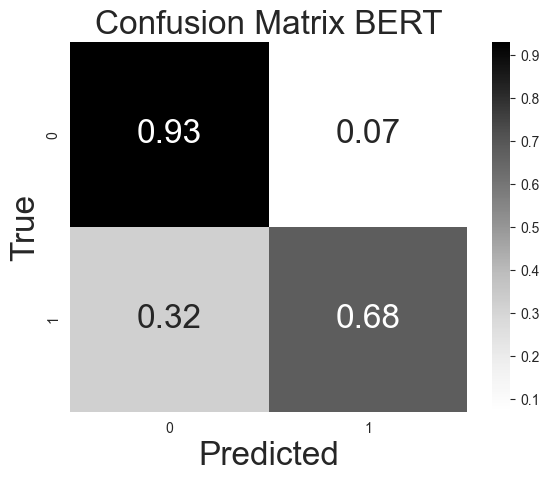}
    \caption{Confusion matrix showing the prediction accuracy of the different models on hate speech detection.}
    \label{fig:confusionmatrix}
\end{figure}

To gain a clearer insight into these outcomes, Figure \ref{fig:confusionmatrix} offers a visual representation illustrating the distribution of accurate and inaccurate predictions generated by our best classification model BERT and SVM. CNN confusion matrix is similar to the SVM one, but due to space limits, we only included the visual representation of SVM. It is important to note that for this classification task, false negatives are more important than false positives. This prioritization stems from the urgency to promptly identify instances where individuals are exposed to harmful content, with the associated benefits outweighing the potential drawbacks.

\section{Discussion and Conclusions}
\label{sec:disc}

Online social networks serve as a valuable repository of digital communication, occasionally becoming a breeding ground for hate speech discourse. It is imperative to address this issue with different models, that rely on large amounts of data to successfully perform. In this work, we presented \textbf{MetaHate}\footnote{Code available here: \url{https://github.com/palomapiot/metahate/}}: a meta-collection encompassing 36 hate speech datasets, representing the first large benchmark corpus on hate speech detection on social media. Additionally, we established a robust baseline, offering future researchers a clear reference for comparison. While we have conformed to the hate speech definitions put forth by the United Nations and previous works, it is important to note that, currently, there is no universally accepted definition of hate speech. Nevertheless, our objective is to move in the direction of establishing such a consensus. Moreover, in enhancing hate speech detection systems, the importance of context and multilingualism has become evident. Current methods often focus on individual English messages, lacking a broader and inclusive view. Future research should aim to capture complete conversation contexts, including different languages, for more accurate and nuanced detection of hate speech online. This work presents another limitation as we adopted a binary classification for hate speech data.

\section{Ethical Statement}
\label{sec:ethical}

Addressing online hate speech involves navigating ethical considerations, particularly in the context of free speech controversies. Our data collection methods encompassed accessing publicly available datasets, obtaining information through submitted forms, and contacting authors via email. Importantly, the collected data strictly lacked any personally identifiable information. The significance of our work extends to potential societal benefits, offering a meticulously curated meta-collection of hate speech data that can enhance the identification of harmful comments on social media platforms. In light of the existence of offensive elements within the meta-collection, it is essential to exercise caution to prevent any potential misapplication for negative purposes, including the propagation of animosity or the targeting of individuals or communities.

Despite our contributions, it is crucial to acknowledge certain limitations. While some datasets are publicly accessible without specific terms and conditions, others necessitate explicit permission from the authors, often requiring individual dataset contracts for MetaHate availability. 

We emphasize that any unintentional biases in the dataset are not intended to cause harm to individuals or target communities; rather, they may be inherent in the original publications. To promote further research in hate speech detection, we share our data and code. However, it is imperative to note that our dataset is released exclusively for research purposes and is not licensed for commercial or malicious use.

\section*{Acknowledgments}

The authors thank the funding from the Horizon Europe research and innovation programme under the Marie Skłodowska-Curie Grant Agreement No. 101073351. The authors also thank the financial support supplied by the Consellería de Cultura, Educación, Formación Profesional e Universidades (accreditation 2019-2022 ED431G/01, ED431B 2022/33) and the European Regional Development Fund, which acknowledges the CITIC Research Center in ICT as a Research Center of the Galician University System and the project PID2022-137061OB-C21 (Ministerio de Ciencia e Innovación supported by the European Regional Development Fund). The authors also thank the funding of project PLEC2021-007662 (MCIN/AEI/10.13039/501100011033, Ministerio de Ciencia e Innovación, Agencia Estatal de Investigación, Plan de Recuperación, Transformación y Resiliencia, Unión Europea-Next Generation EU).

\bigskip

\bibliography{metahate}
\setlength{\tabcolsep}{4pt}

\footnotesize
\bottomcaption{Hate Speech English datasets. Datasets in italic and gray colour are not included in MetaHate.}

\tablefirsthead{\toprule \textbf{Dataset} & \textbf{Actual Size} & \textbf{Source} & \textbf{Classification Type} & \textbf{Labels} &  \textbf{Baseline} & \textbf{Creation Strategy} & \textbf{Available} \\ \midrule}

\tablehead{
	\multicolumn{2}{l}
	{{\bfseries  Continued from previous page}} \\
	\toprule
	\textbf{Dataset} & \textbf{Actual Size} & \textbf{Source} & \textbf{Classification Type} & \textbf{Labels} &  \textbf{Baseline} & \textbf{Creation Strategy} & \textbf{Available}\\}

\tabletail{
	\midrule \multicolumn{2}{l}{{Continued on next page}} \\ \midrule}
\tablelasttail{
	\bottomrule}

\onecolumn
\appendix
\section{Appendix}
\begin{xtabular}{p{2.2cm} p{1.1cm} p{1.45cm} p{1.7cm} p{3cm} p{1.2cm} p{2.9cm} p{1.4cm}}
   Online Harassment 2017 & 19,838 & Twitter & Binary & hate, no hate & - & Lexicon, word structures, hashtags & Via email \\ 
    \midrule
    \textcolor{gray}{\textit{AMI 2018}} & \textcolor{gray}{\textit{3,251}} & \textcolor{gray}{\textit{Twitter}} & \textcolor{gray}{\textit{Binary}} & \textcolor{gray}{\textit{misogyny, no misogyny}}  & \textcolor{gray}{\textit{SVM}} & \textcolor{gray}{\textit{Keywords, account monitoring}} & \textcolor{gray}{\textit{Protected with password}} \\
    \midrule
    \textcolor{gray}{\textit{Cyberbullying Personality}} & \textcolor{gray}{\textit{3,987}} & \textcolor{gray}{\textit{Twitter}} & \textcolor{gray}{\textit{Binary}} & \textcolor{gray}{\textit{cyberbullying, no cyberbullying}} & \textcolor{gray}{\textit{Random Forest}} & \textcolor{gray}{\textit{Hashtag}} & \textcolor{gray}{\textit{No}} \\
    \midrule
    \textcolor{gray}{\textit{Hate Lingo 2018}} & \textcolor{gray}{\textit{148,256}} & \textcolor{gray}{\textit{Twitter}} & \textcolor{gray}{\textit{Binary}} & \textcolor{gray}{\textit{directed hate, generalized hate}} & \textcolor{gray}{\textit{-}} & \textcolor{gray}{\textit{Lexicon, hashtags, dataset sampling, random sampling}} & \textcolor{gray}{\textit{Only Tweet IDs}} \\
    \midrule
    Hateval 2019 & 12,747 & Twitter & Binary & hate, no hate & SVM & Lexicon, hashtags, user monitoring & Yes \\
    \midrule
    OLID 2019 & 14,052 & Twitter & Binary & offensive, not offensive & CNN & Keywords, phrase structures & Yes \\
    \midrule
    \textcolor{gray}{\textit{SWAD 2020}} & \textcolor{gray}{\textit{2,569}} & \textcolor{gray}{\textit{Twitter}} & \textcolor{gray}{\textit{Binary}} & \textcolor{gray}{\textit{hate, no hate}}  & \textcolor{gray}{\textit{LSVC}} & \textcolor{gray}{\textit{OLID subset using keywords}} & \textcolor{gray}{\textit{Yes}} \\ 
    \midrule
    US 2020 Elections & 2,999 & Twitter & Binary & hate, no hate  & BERT & Keywords, hashtags & Yes \\
    \midrule
    ``Call me sexist but'' 2021 & 3,058 & Twitter & Binary & sexism, no sexism  & BERT & Sexism scales, phrase ``call me sexism, but'', datasets sampling & Via registration \\
    \midrule
    HASOC 2019-2021 & 6,981 & Twitter, Facebook & Binary & hate, no hate & LSTM & Heuristics, topics, keywords and hashtags & Only 2019 \\
    \midrule
    A Curated Hate Speech Dataset 2023 & 560,385 & Twitter, Facebook, Wikipedia, etc. & Binary & hate, no hate & - & Not explained & Yes \\
    \midrule
    Hate Speech B 2016 & 6,909 & Twitter & Multiclass & racism, sexism, both, none & n-grams & \citep{Waseem2016} sample & Via email \\
    \midrule
    Hate Speech A 2016 & 16,849 & Twitter & Multiclass & racism, sexism, none & Logistic Regression & Lexicon & Via email \\
    \midrule
    \textcolor{gray}{\textit{Mean Birds 2017}} & \textcolor{gray}{\textit{9,484}} & \textcolor{gray}{\textit{Twitter}} & \textcolor{gray}{\textit{Multiclass}} & \textcolor{gray}{\textit{normal, spammer, aggressor, bully}} & \textcolor{gray}{\textit{Random Forest}} & \textcolor{gray}{\textit{Hashtags}} & \textcolor{gray}{\textit{Only Tweet IDs}}\\ 
    \midrule
    \textcolor{gray}{\textit{Ambivalent Sexism 2017}} & \textcolor{gray}{\textit{22,142}} & \textcolor{gray}{\textit{Twitter}} & \textcolor{gray}{\textit{Multiclass}} & \textcolor{gray}{\textit{hostile, benevolent, others}} & \textcolor{gray}{\textit{FastText}} & \textcolor{gray}{\textit{Keywords, hashtags, phrase structures}} & \textcolor{gray}{\textit{Only Tweet IDs}} \\
    \midrule
    Hate Offensive 2017 & 24,783 & Twitter & Multiclass & hate speech, offensive, neither & Logistic Regression & Lexicon & Yes \\ 
    \midrule
    TRAC1 2018 & 14,537 & Twitter, Facebook & Multiclass & very aggressive, a bit aggressive, not aggressive & - & Hashtags, controversial topics & Via form \\
    \midrule
    \textcolor{gray}{\textit{Harassment Corpus 2018}} & \textcolor{gray}{\textit{25,000}} & \textcolor{gray}{\textit{Twitter}} & \textcolor{gray}{\textit{Multiclass}} & \textcolor{gray}{\textit{sexual, racial, appearance-related, intellectual, political}} & \textcolor{gray}{\textit{-}} & \textcolor{gray}{\textit{Lexicon}} & \textcolor{gray}{\textit{No}} \\
    \midrule
    ENCASE 2018 & 91,950 & Twitter & Multiclass & abusive, normal, spam, hateful & - & Random sample, lexicon & Yes \\ 
    \midrule
    MLMA 2019 & 5,593 & Twitter & Multiclass, multilabel & abusive, hateful, offensive, disrespectful, fearful, normal & biLSTM & Keywords & Yes \\
    \midrule
    \#MeTooMA 2020 & 9,889 & Twitter & Multiclass & directed hate, generalized hate, sarcasm, allegation, justification, refutation, support, oppose & - & Lexicon, filtering by country & Via email \\
    \midrule
    HateXplain 2020 & 20,109 & Twitter, Gab & Multiclass & hate, offensive, normal  & BERT & Lexicon & Yes \\
    \midrule
    Hate Speech Data 2017 & 6,157 & \textcolor{gray}{\textit{Twitter}}, Whisper & Multiclass & race, behaviour, physical, sexual orientation, class, gender, ethnicity, disability, religion, other & - & Sentence structures & Only Whisper \\
    \midrule
    Hateful Tweets 2022 & 1,141 & Twitter & Multiclass & hate, justification, attacks author, additional hate  & NA & Existing datasets extension & Via email \\
    \midrule
    Multiclass Hate Speech 2022 & 68,597 & Twitter & Multiclass & hate, offensive, normal  & Megatron (BERT) & Keywords, hashtags & Via email \\
    \midrule
    Measuring Hate Speech 2020/2022 & 39,565 & Twitter, Reddit, YouTube & Probability & higher is more hate, lower less  & RoBERTa & Random Sample & Yes \\
    \midrule
    BullyDetect 2018 & 6,562 & Reddit & Binary & cyberbullying, no cyberbullying & Random Forest & Not explained & Yes \\
    \midrule
    Intervene Hate 2019 & 45,170 & Reddit, Gab & Binary & hate, no hate  & CNN/RNN & Toxic subreddits, keywords & Yes \\
    \midrule
    Slur Corpus 2020 & 39,960 & Reddit & Multiclass & derogatory, non derogatory, homonym, appropriation, noise  & - & \citep{baumgartner2020-pushshift} keywords filtering & Yes \\
    \midrule
    CAD 2021 & 23,060 & Reddit & Multiclass, multilabel & identity directed abuse, affiliation directed abuse, person directed abuse, counter speech, neutral  & BERT &  Community-based sampling & Yes \\
    \midrule
    ETHOS 2022 & 998 & Reddit, YouTube & Probability & 1 hate, 0 no hate & DistilBERT & Subreddits, \url{hatebusters.org} & Yes \\
    \midrule
    Hate in Online News Media 2018 & 3,214 & Facebook, YouTube & Binary & hate, neutral & SVM & Filtering by an online news and media company & Yes \\
    \midrule
    Supremacist 2018 & 10,534 & Stormfront & Binary & hate, no hate & LSTM & Random sample & Yes \\ 
    \midrule
    The Gab Hate Corpus 2022 & 27,434 & Gab & Binary & assault on human dignity, not assault on human dignity  & BERT &  Random sample & Yes \\
    \midrule
    HateComments 2023 & 2,070 & YouTube, BitChute & Binary & hate, no hate & BERT & Manual selection from list of categories & Yes \\
    \midrule
    TRAC2 2020 & 5,329 & YouTube & Multiclass & very aggressive, a bit aggressive, not aggressive & - & Selected topics & Via form \\
    \midrule
    Toxic Spans 2021 & 10,621 & Civil Comments & Spans & Span positions  & BERT & Civil Comments labelled as toxic & Yes \\
    \midrule
    \textcolor{gray}{\textit{Dynamic Hate 2021}} & \textcolor{gray}{\textit{41,135}} & \textcolor{gray}{\textit{Synthetic}} & \textcolor{gray}{\textit{Binary}} & \textcolor{gray}{\textit{hate, no hate}} & \textcolor{gray}{\textit{RoBERTa}} & \textcolor{gray}{\textit{Data generation}} & \textcolor{gray}{\textit{Yes}} \\
    \midrule
    \textcolor{gray}{\textit{CONAN 2019-2022}} & \textcolor{gray}{\textit{8,883}} & \textcolor{gray}{\textit{Semi-Synthetic}} & \textcolor{gray}{\textit{Multiclass}} & \textcolor{gray}{\textit{disabled, jews, LGBT+, migrants, muslims, people of color, women}} & \textcolor{gray}{\textit{-}} & \textcolor{gray}{\textit{LMs}} & \textcolor{gray}{\textit{Yes}} \\
    \midrule
    Ex Machina 2016 & 115,705 & Wikipedia & Binary & attack, no attack & MLP & Random sample, blocked users & Yes \\
    \midrule
    Context Toxicity 2020 & 19,842 & Wikipedia & Binary & toxic, no toxic & BERT & Not explained & Yes \\
    \midrule
    \textbf{MetaHate} & \textbf{1,226,202} & \textbf{Social media} & \textbf{Binary} & \textbf{hate, no hate} & \textbf{BERT} & \textbf{Meta collection} & \textbf{Yes} \\
\end{xtabular}
\label{tab:datasets}

\begin{table*}[!htb]
\centering
\begin{adjustbox}{width=1\textwidth}
\begin{tabular}{p{1.6cm} p{1.2cm} p{2.5cm} p{1cm} p{1.5cm} l p{2.2cm} p{3.8cm} l}
\toprule
\textbf{Author} & \textbf{Platform} & \textbf{Link} & \textbf{Lang.} & \textbf{Actual Size} & \textbf{Source} & \textbf{Classification Type} & \textbf{Labels}  & \textbf{Available} \\
\midrule
Roshan Sharma / Ali Toosi & Hugging Face &  \url{https://t.ly/v9tin} & EN & 29,530 & Twitter & Binary & hate, no hate & Yes \\
\midrule
Munki Albright & Kaggle & \url{https://t.ly/ZO_Cx} & EN & 18,208 & Twitter & Multiclass & suspicious, cyberbullying, hate, suicidal & Yes \\ 
\midrule
SR & Kaggle & \url{https://t.ly/XpBNt} & EN & 159,571 & Twitter & Multiclass & malignant, highly malignant, rude, threat, abuse, loathe & Yes \\ 
\midrule
Jigsaw & Kaggle & \url{https://t.ly/v6IkS} & EN & 223,549 & Wikipedia & Multiclass, multilabel & toxic, severe toxic, obscene, threat, insult, identity hate & Yes \\ 
\bottomrule
\end{tabular}
\end{adjustbox}
\caption{Hate Speech Datasets from different platforms.}
\label{tab:datasets_repos}
\end{table*}

\normalsize
\twocolumn

\end{document}